\documentclass[final,acmtog,screen,nonacm]{acmart}

\AtBeginDocument{%
  }

\usepackage{fix-cm}
\usepackage{amsfonts}
\usepackage{amsmath}
\usepackage{amsthm}
\usepackage{latexsym}
\usepackage{bm}
\usepackage{multirow}
\usepackage{graphicx}
\usepackage{enumitem}
\usepackage{array}
\usepackage{wrapfig}
\usepackage{xcolor}
\usepackage[ruled, linesnumbered]{algorithm2e}
\usepackage{nicefrac}
\usepackage{microtype}
\usepackage{pifont}
\usepackage{subcaption}
\usepackage{algpseudocode}
\usepackage{siunitx}

\newcommand{\colref}[3]{\hyperref[#2]{#1~\ref*{#2}{#3}}}
\newcommand{\figref}[1]{\colref{Fig.}{#1}{}}
\newcommand{\figrefA}[2]{\colref{Fig.}{#1}{#2}}
\newcommand{\secref}[1]{\colref{Section}{#1}{}}

\newcommand{\tabref}[1]{\colref{Table}{#1}{}}

\newcommand{\algoref}[1]{\colref{Algorithm}{#1}{}}

\graphicspath{{figures_CEA/}}

\begin{document}

\begin{teaserfigure}
    \centering
    \includegraphics[width=0.99\linewidth,trim= 0.0in 0.0in 0.0in 0.0in, clip]{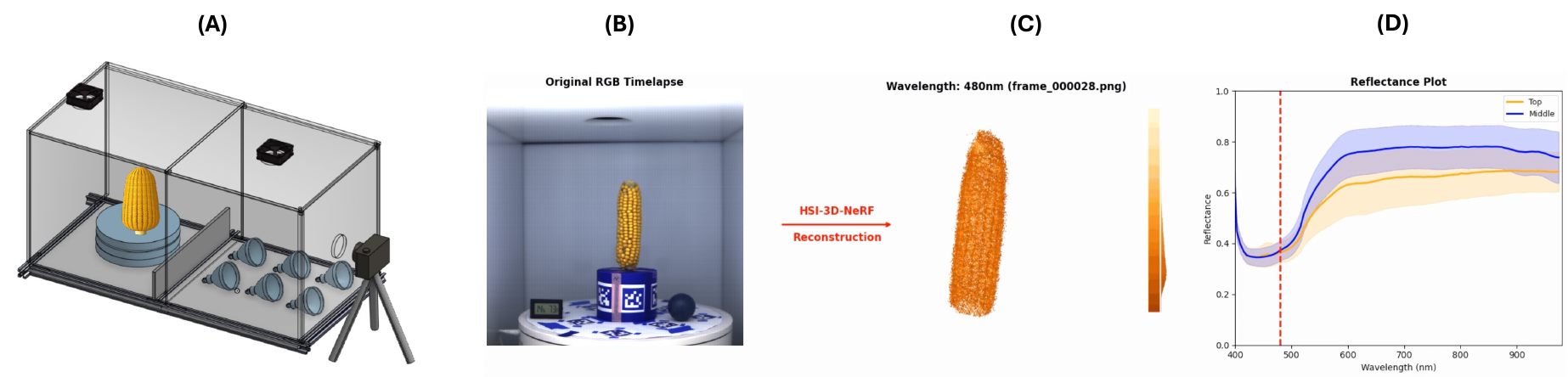}
    \caption{Schematic of the stationary camera imaging system for hyperspectral NeRF-based point cloud reconstruction in high-throughput plant phenotyping. In this setup, each plant is conveyed to a rotating turntable marked against a Teflon (PTFE) background. Over a full rotation, a tripod-mounted stationary camera captures high-resolution images that serve as input for HS-3D-NeRF techniques to generate 3D reconstructions. This streamlined approach eliminates the need for complex moving-camera rigs, aligning with the objectives of efficient, scalable agricultural imaging. The right shows the different PCD reconstructions using HS-3D-NeRF and spectral signatures on the circled regions. }
    \Description[Teaser]{Overview of HS-3D-NeRF}
    \label{fig:Teaser}
\end{teaserfigure}

\title[Multi-Channel NeRF for Hyperspectral 3D Reconstruction]{HS-3D-NeRF: 3D Surface and Hyperspectral Reconstruction From Stationary Hyperspectral Images Using Multi-Channel NeRFs}

\author{Kibon Ku}
\email{kibona9@iastate.edu}
\authornotemark[1]
\author{Talukder Z. Jubery}
\email{znjubery@iastate.edu}
\authornote{Equal contribution}
\author{Adarsh Krishnamurthy}
\authornotemark[2]
\email{adarsh@iastate.edu}
\author{Baskar Ganapathysubramanian}
\authornote{Corresponding authors}
\email{baskarg@iastate.edu}
\affiliation{\institution{Iowa State University} \city{Ames} \state{Iowa} \country{USA}}

\renewcommand{\shortauthors}{Ku and Jubery et al., 2025}

\begin{abstract}
Advances in hyperspectral imaging (HSI) and 3D reconstruction have enabled accurate, high-throughput characterization of agricultural produce quality and plant phenotypes, both essential for advancing agricultural sustainability and breeding programs. HSI captures detailed biochemical features of produce, while 3D geometric data substantially improves morphological analysis. However, integrating these two modalities at scale remains challenging, as conventional approaches involve complex hardware setups incompatible with automated phenotyping systems. Recent advances in neural radiance fields (NeRF) offer computationally efficient 3D reconstruction but typically require moving-camera setups, limiting throughput and reproducibility in standard indoor agricultural environments. To address these challenges, we introduce HSI-SC-NeRF, a stationary-camera multi-channel NeRF framework for high-throughput hyperspectral 3D reconstruction targeting postharvest inspection of agricultural produce. Multi-view hyperspectral data is captured using a stationary camera while the object rotates within a custom-built Teflon imaging chamber providing diffuse, uniform illumination. Object poses are estimated via ArUco calibration markers and transformed to the camera frame of reference through simulated pose transformations, enabling standard NeRF training on stationary-camera data. A multi-channel NeRF formulation optimizes reconstruction across all hyperspectral bands jointly using a composite spectral loss, supported by a two-stage training protocol that decouples geometric initialization from radiometric refinement. Experiments on three agricultural produce samples demonstrate high spatial reconstruction accuracy and strong spectral fidelity across the visible and near-infrared spectrum, confirming the suitability of HSI-SC-NeRF for integration into automated agricultural workflows.
\end{abstract}

\ccsdesc[500]{Applied computing~Agriculture}
\ccsdesc[500]{Computing methodologies~Hyperspectral imaging}
\ccsdesc[300]{Computing methodologies~Computational photography}
\ccsdesc[100]{Computing methodologies~Reflectance modeling}

\keywords{Hyperspectral Imaging, Multi-channel NeRFs, Hyperspectral 3D Reconstruction}


\maketitle

\section{Introduction}
\label{sec:intro}

High-throughput and precise characterization of plant phenotypes and produce quality is critical for advancing food security, enhancing nutritional quality, and promoting agricultural sustainability~\citep{gupta2024ai, sarkar2023cyber}. Advanced imaging techniques that integrate spatial and spectral information enable simultaneous morphological and biochemical trait analysis, accelerating breeding programs and improving supply chain monitoring. Hyperspectral imaging (HSI), which captures reflectance across hundreds of narrow spectral bands spanning the visible to near-infrared spectrum (400--1000~\unit{nm}), has proven invaluable in agriculture due to its sensitivity to physiological indicators such as chlorophyll content, water stress, and biochemical composition~\citep{song2024fruit, li2025bruise}.

\begin{figure}[t!]
    \centering
    \includegraphics[width=0.99\linewidth]{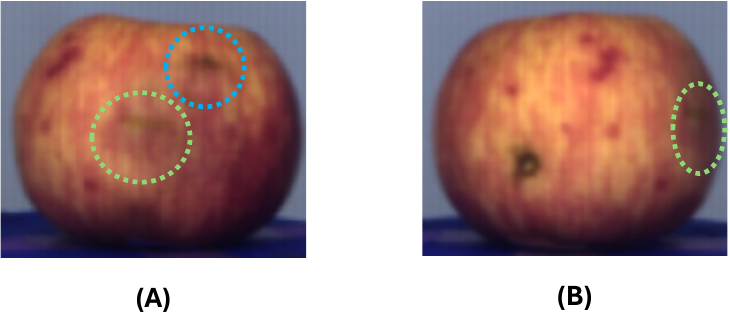}
    \caption{Motivation for hyperspectral imaging (HSI) in 3D.
    Two views of the same apple show viewpoint-dependent bruise visibility: one ROI is visible in view (A) but occluded in view (B), motivating HSI with 3D reconstruction for more complete surface assessment.}    
    \label{fig:motivation_hsi_3d}
\end{figure}

Traditional HSI methods generally produce 2D hyperspectral data, limiting their ability to accurately characterize 3D plant architecture or produce morphology. As illustrated in \figref{fig:motivation_hsi_3d}, surface curvature and self-occlusion can cause a defect region to be visible from one viewing angle but partially or fully hidden from another, resulting in missed detections and biased severity estimates under single-view analysis. This limitation has motivated significant research into integrating hyperspectral data with 3D structural reconstruction using techniques such as Structure-from-Motion (SfM), Multi-View Stereo (MVS)~\citep{eltner2020structure, chen2019point}, or active depth sensing technologies such as LiDAR~\citep{wang2020lidar, guo2023integrated}. However, these approaches often involve complex hardware setups with high operational costs and demanding calibration procedures, limiting their scalability in high-throughput agricultural workflows~\citep{andujar2018three, tang2022benefits, paturkar2021effect, lu2023bird}.

Recent advances in neural implicit representations, particularly neural radiance fields (NeRF), offer robust and scalable alternatives for 3D reconstruction~\citep{mildenhall2021nerf}. NeRFs model volumetric scenes by learning radiance and density distributions from multi-view images without explicit geometric constraints, and can replicate complex view-dependent effects such as translucency and occlusion that are common in agricultural subjects~\citep{cuevas2020segmentation, feng20233d, sarkar2023cyber}. The integration of HSI into the NeRF framework further enables simultaneous recovery of geometry and spectral reflectance information at each spatial point~\citep{chen2024hyperspectral}.

Despite these advantages, conventional NeRF implementations require camera movement around stationary objects to capture multiple viewpoints, posing logistical challenges in high-throughput phenotyping environments where automated conveyors and rotating pedestals are standard. Multi-camera arrays and LiDAR scanners can provide the necessary viewpoint coverage but remain cost-prohibitive and operationally complex~\citep{medic2023challenges}, while reconstruction quality remains sensitive to lighting variation, occlusion, and motion-induced blur. A stationary-camera NeRF approach would substantially improve scalability and throughput in both phenotyping and postharvest inspection pipelines.

To address these limitations, we introduce \textbf{HSI-SC-NeRF}, a stationary-camera hyperspectral multi-channel NeRF reconstruction framework developed for high-throughput agricultural imaging. Rather than moving the camera, the object rotates on a motorized turntable within a custom-built Teflon imaging chamber that provides diffuse, uniform illumination. COLMAP-based pose estimation using ArUco calibration markers, combined with a simulated pose transformation to the camera frame of reference, enables standard NeRF training on the resulting stationary-camera data. A multi-channel NeRF formulation then optimizes reconstruction jointly across all hyperspectral bands via a composite spectral loss and a two-stage training protocol that separates geometric initialization from radiometric refinement.

The main contributions of this work are:
\begin{enumerate}[label=(\arabic*)]
    \item A stationary-camera NeRF pipeline with ArUco-based pose estimation and simulated pose transformation, enabling multi-view hyperspectral acquisition without moving the camera.
    \item A custom Teflon imaging chamber providing diffuse, uniform illumination for consistent and repeatable hyperspectral data acquisition.
    \item A multi-channel NeRF formulation with a composite spectral loss and two-stage training protocol that jointly optimizes geometry and spectral fidelity across all hyperspectral bands.
    \item Quantitative validation of spatial reconstruction accuracy and wavelength-resolved spectral fidelity on three agriculturally relevant produce samples.
\end{enumerate}

Taken together, these contributions demonstrate the feasibility of integrating hyperspectral imaging and 3D reconstruction into automated phenotyping and postharvest quality control workflows. The framework is particularly well-suited to high-value crops, and the multi-channel NeRF formulation is of independent interest to the graphics community as a method for capturing physically meaningful spectral data~\citep{boss2021nerd, li2024spectralnerf}. 


\begin{figure*}[t!]
    \centering
    \includegraphics[width=0.9\linewidth]{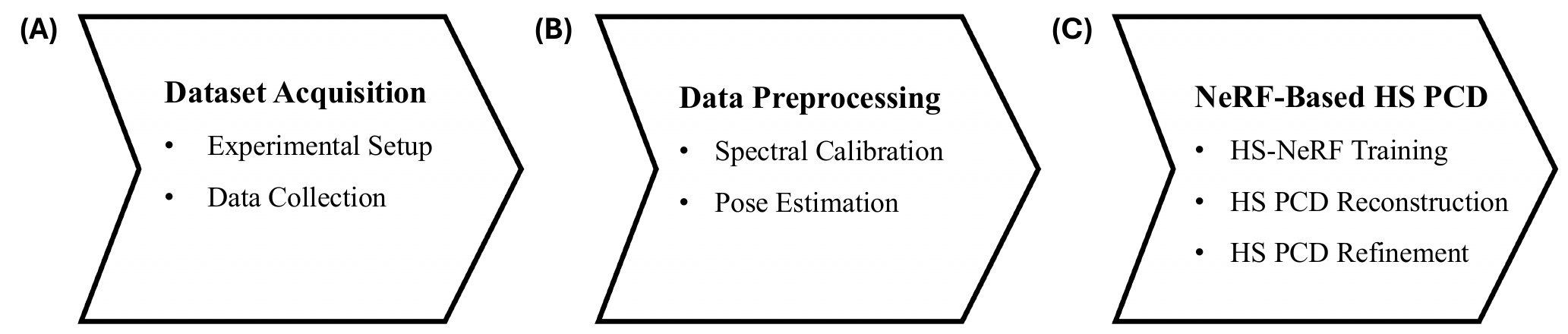}
    \includegraphics[width=1\linewidth,trim={0.0in 0.0in 0.0in .3in},clip]{graphfical_abstract.pdf}
    \caption{Workflow of the HSI-SC-NeRF pipeline. The process consists of three main steps: (A) Dataset Acquisition, where the experimental environment is set up and multi-view hyperspectral image data is collected using a stationary camera; (B) Data Preprocessing, involving white-reference spectral calibration and COLMAP-based pose estimation to ensure radiometric and geometric consistency; and (C) NeRF-Based HS PCD, where a multi-channel NeRF model is trained for scene representation, followed by hyperspectral point cloud reconstruction and refinement to generate high-quality 3D hyperspectral point clouds. The graphical abstract (bottom) illustrates the end-to-end pipeline: the imaging chamber with rotating turntable (A), an RGB timelapse frame from the acquisition session (B), the resulting hyperspectral point cloud at 480~\unit{nm} (C), and per-region reflectance spectra extracted from the reconstructed point cloud (D).}
    \label{fig:Workflow}
\end{figure*}

\section{Related Work}
\label{Sec:Related}

\noindent\paragraph{Hyperspectral Imaging in Agriculture and 3D Reconstruction}
Hyperspectral imaging (HSI) has significantly advanced agricultural applications over the past two decades by enabling non-destructive assessment of plant and produce quality~\citep{gowen2007hyperspectral}. Capturing hundreds of narrow spectral bands spanning the visible to near-infrared spectrum (400--1000~\unit{nm}), HSI reveals biochemical indicators such as chlorophyll content, water status, and ripeness~\citep{mahlein2013development}. Early approaches integrated HSI with 3D structural data using Structure-from-Motion (SfM), Multi-View Stereo (MVS), LiDAR, or structured light scanners~\citep{paulus2014plant}. Although these methods provide accurate spectral and geometric reconstruction, their practical application is constrained by complex hardware, calibration difficulties, and low throughput. Gantry- and turntable-mounted setups have sought to simplify multi-view acquisition, but remain cumbersome due to self-occlusion, inconsistent viewpoint synchronization, and complex data fusion. These limitations underscore the need for scalable techniques capable of simultaneously capturing rich spectral and geometric data in high-throughput agricultural settings.

\noindent\paragraph{Neural Radiance Fields for Spectral 3D Reconstruction}
Neural Radiance Fields (NeRF) have substantially advanced 3D reconstruction by modeling scenes as continuous volumetric functions learned from multi-view images~\citep{mildenhall2021nerf}. NeRF approaches have recently been extended beyond RGB to include hyperspectral data, with studies demonstrating robust spectral radiance field learning from limited viewpoints~\citep{chen2024hyperspectral}. Spec-NeRF, for example, reconstructs spectral radiance fields from RGB cameras with interchangeable filters, reducing hardware complexity~\citep{li2024spec}. In agricultural imaging, NeRF-based methods have proven effective for detailed crop reconstruction, accurately estimating traits such as leaf area and fruit volume~\citep{meyer2024fruitnerf}. Despite these advances, practical limitations persist: the extensive viewpoint coverage required, lengthy capture times, and intensive computational demands continue to hinder deployment in high-throughput contexts.

\noindent\paragraph{Limitations of Moving-Camera NeRF in High-Throughput Settings}
Traditional NeRF setups require numerous images from varied viewpoints, necessitating camera motion around stationary objects or multi-camera rigs~\citep{gao2023mc}. Such setups restrict throughput in agricultural environments due to mechanical complexity, synchronization challenges, and lengthy calibration processes. Maintaining consistent lighting and spectral calibration across multiple viewpoints further exacerbates these issues, leaving moving-camera NeRF implementations predominantly confined to controlled experimental settings and unsuitable for fast-paced industrial or agricultural inspection.

\noindent\paragraph{HSI-SC-NeRF}
The proposed methodology addresses these limitations through two complementary advances. First, hyperspectral data acquisition is performed with a stationary camera while the object rotates on a motorized turntable within a custom Teflon imaging chamber, simplifying multi-view capture and eliminating the mechanical complexities of traditional setups. Second, the NeRF formulation is extended to predict multi-channel hyperspectral outputs, enabling simultaneous recovery of detailed geometry and spectral reflectance under consistent illumination. Together, these advances improve scalability and support advanced post-reconstruction analyses such as early disease detection, nutrient assessment, and maturity estimation~\citep{nguyen2021early}.

\section{Materials and Methods}
\label{sec:methods}

The HSI-SC-NeRF pipeline, illustrated in \figref{fig:Workflow}, proceeds in three sequential stages. In the \textbf{Dataset Acquisition} stage (\secref{sec:setup}), an object is rotated on a motorized turntable within a custom Teflon imaging chamber while a stationary SPECIM IQ hyperspectral camera captures one frame every two minutes over a full rotation, yielding 60 frames per object across 204 spectral bands; ArUco markers on the turntable and a 3D-printed cylindrical reference container provide persistent features for pose estimation. In the \textbf{Data Preprocessing} stage (\secref{subsec:spectral_calibration}), the raw hyperspectral cubes undergo white-reference spectral calibration to correct for wavelength-dependent illumination bias, followed by COLMAP-based pose estimation whose object-frame outputs are transformed to the camera frame of reference via simulated pose transformations. In the final \textbf{NeRF-Based Hyperspectral Point Cloud Reconstruction} stage (\secref{subsec:nerf}), a multi-channel NeRF supervised by a composite spectral loss is trained using a two-stage protocol that separates full-frame geometric initialization from masked radiometric fine-tuning, and the trained model is used to reconstruct and refine a dense 3D hyperspectral point cloud.

\subsection{Experimental Setup and Data Collection}
\label{sec:setup}

\begin{figure*}[t!]
    \centering
    \includegraphics[width=0.9\linewidth]{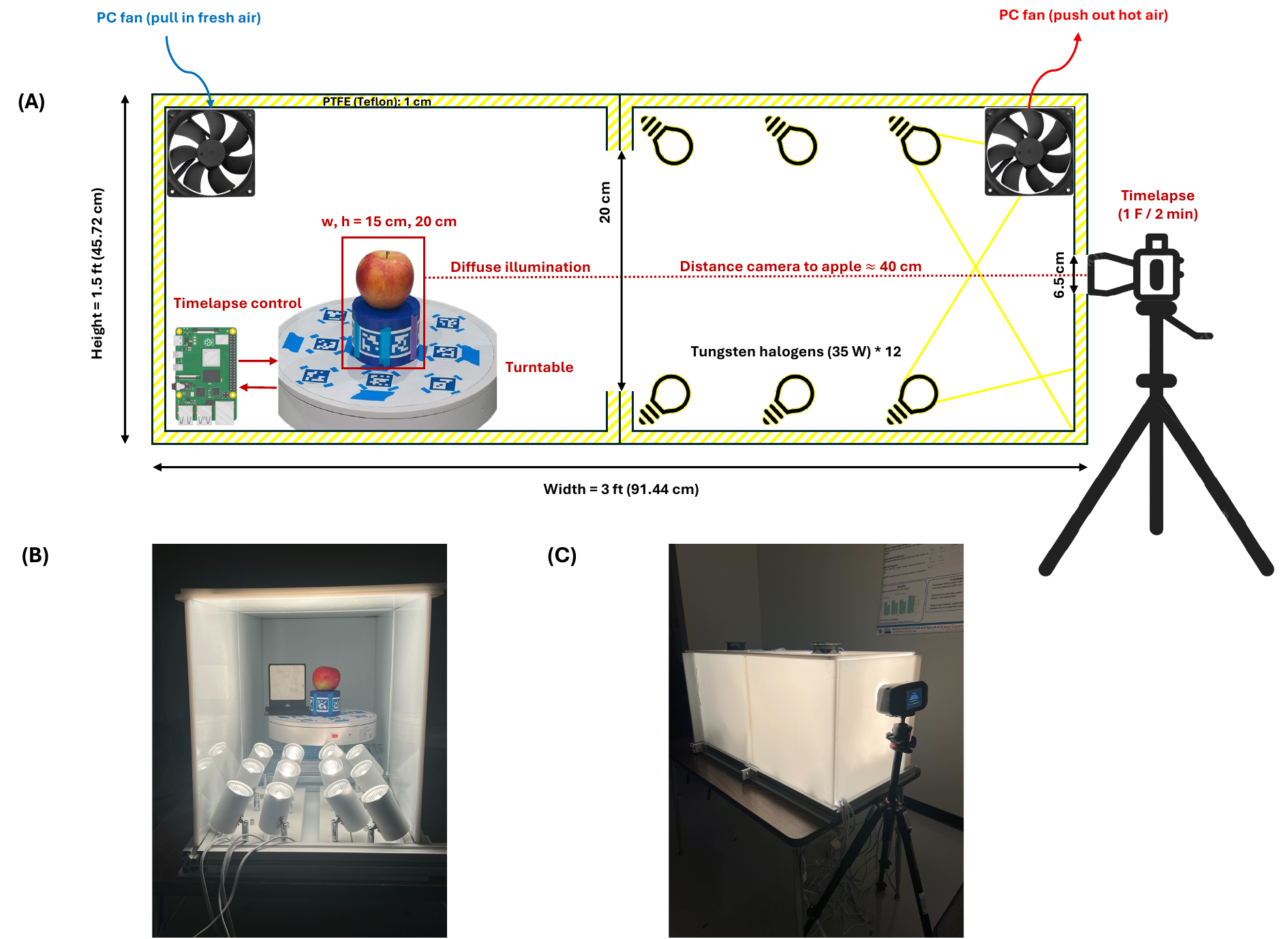}
    \caption{Teflon studio chamber for uniform hyperspectral illumination. (A) The chamber is designed from PTFE (Teflon) sheets to provide diffuse illumination. (B) Inside the chamber and (C) the outside.}
    \label{fig:Exp_Setup}
\end{figure*}

The experimental setup (\figref{fig:Exp_Setup}) was designed around the SPECIM IQ camera (SPECIM, Spectral Imaging Ltd., Oulu, Finland; \url{https://www.specim.fi/specim-iq/}), a portable push-broom hyperspectral imaging system capturing 204 spectral bands from 397 to 1003~\unit{nm} at a spectral resolution of approximately 7~\unit{nm} and a spatial resolution of $512 \times 512$ pixels in band-interleaved-by-line (BIL) format. The camera was operated with no binning, a frame rate of 100~\unit{fps}, and an integration time of 10~\unit{ms}. It was mounted on a tripod, oriented orthogonally to the object, and remained stationary throughout all acquisitions. The object was placed on a motorized rotating turntable inside a custom Teflon (PTFE) imaging chamber lined with twelve 35~W tungsten halogen lamps, whose reflections off the PTFE walls produce the diffuse, spatially uniform illumination required for consistent hyperspectral reflectance measurement. The Teflon background improved foreground segmentation, and the entire setup was placed on a vibration-isolated surface to minimize motion artifacts.

For pose estimation, $5\times5$ ArUco markers were deployed in two configurations. Six markers were affixed to a custom 3D-printed cylindrical container (diameter 0.09~\unit{m}, height 0.07~\unit{m}) designed in OnShape (\url{https://www.onshape.com/}) and fabricated on an Ultimaker S3 FDM printer. The container's smooth, uniformly colored surface minimized blending artifacts against the Teflon background and provided a constant radius that maintained continuous marker visibility throughout rotation, addressing geometric inconsistencies common in standard plant pots. Its dimensions were measured with a digital caliper to serve as a metric reference. An additional eight markers were arranged on a circular paper disc mounted to the turntable. Together, these markers provided persistent features for COLMAP to compute camera extrinsic parameters, which is essential in a stationary-camera setup where all parallax and viewpoint variation arises solely from object and marker movement.

The dataset comprised three objects of varying geometry and structural complexity: an apple, a pear, and a maize ear. Each object was positioned individually on the turntable, whose rotation was controlled via a Python script to ensure consistent angular intervals and full surface coverage. A time-lapse acquisition protocol captured one hyperspectral frame every two minutes for a total of 60 frames per object, with each session lasting approximately two hours and yielding 204-band hyperspectral cubes spanning 397.32 to 1003.58~\unit{nm}. Acquisition speeds can be significantly increased in production settings through faster turntable rotation and reduced inter-frame intervals.

\subsection{Data Processing}
\label{subsec:spectral_calibration}

\subsubsection*{White Reference-Based Spectral Calibration}

\begin{figure*}[t!]
    \centering
    \includegraphics[width=0.9\linewidth]{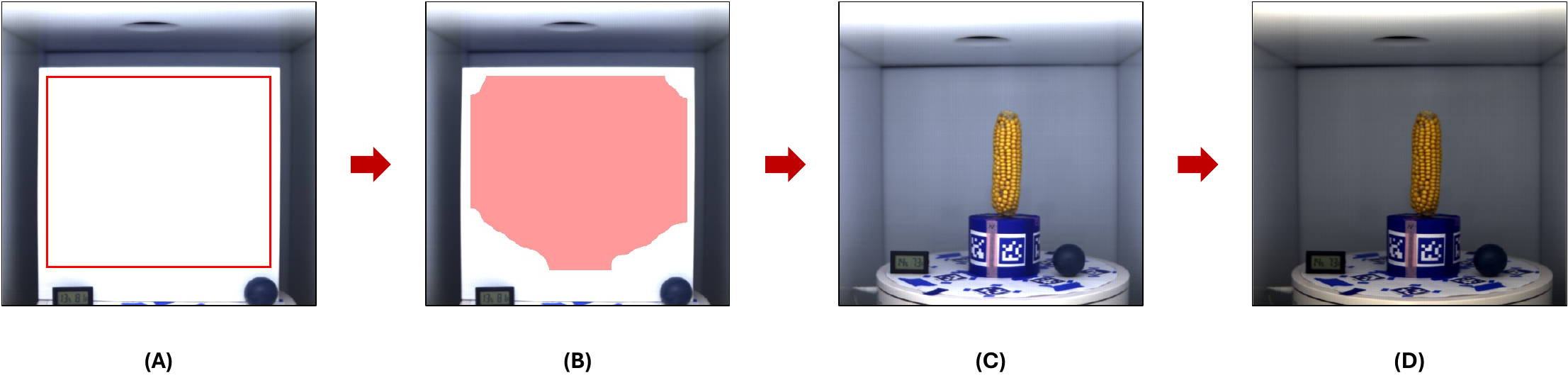}
    \caption{White reference (WR)--based spectral calibration. (A) Coarse WR region of interest (ROI) extraction. (B) Automatically generated central WR mask after 70th-percentile-based spatial deviation filtering. (C) Pseudo-RGB rendering of the raw hyperspectral cube before calibration. (D) Final spectrally calibrated result after WR-based normalization and reflectance clipping.}
    \label{fig:WR_Calibration}
\end{figure*}

To improve spectral consistency under controlled indoor imaging and to mitigate wave\-length-dependent illumination and sensor-response bias, we apply a white reference (WR)--based spectral calibration prior to reconstruction (\figref{fig:WR_Calibration}). The complete procedure is provided in \algoref{alg:wr_calib}. Let $I(x,y,\lambda)$ denote the captured hyperspectral cube with $L$ spectral bands. The goal is to estimate a stable WR spectrum acquired under identical conditions and use it for band-wise normalization.

WR pixels are first extracted from a coarse region of interest (ROI), $\Omega_{\mathrm{WR}}$, selected on a dedicated WR tarp capture. Although the tarp is nominally uniform, spatial non-uniformity typically appears near the boundaries due to illumination falloff and edge effects. To quantify this instability, we compute a per-pixel relative deviation map within $\Omega_{\mathrm{WR}}$:
\begin{equation}
D(x,y) = \frac{1}{L}\sum_{\lambda} \left|\frac{I_{\mathrm{WR}}(x,y,\lambda)-\mu_{\mathrm{WR}}(\lambda)}{\mu_{\mathrm{WR}}(\lambda)}\right|,
\end{equation}
where $I_{\mathrm{WR}}(x,y,\lambda)$ denotes WR pixels and $\mu_{\mathrm{WR}}(\lambda)$ is a preliminary mean WR spectrum computed over the coarse ROI. A refined WR mask is then generated by retaining only pixels satisfying
\begin{equation}
D(x,y) \le \mathrm{Percentile}_{p}(D),
\end{equation}
followed by morphological filtering and largest connected component selection to remove fragmented regions and boundary artifacts. This percentile-based selection yields a spatially contiguous central WR region that preferentially excludes illumination-unstable edge pixels while remaining aligned with the object-centered acquisition geometry (\figref{fig:WR_Calibration}).

The WR mean spectrum is recomputed using only the refined mask $\Omega_{\mathrm{WR}}^{\ast}$:
\begin{equation}
\mu_{\mathrm{WR}}(\lambda) = \frac{1}{N}\sum_{(x,y)\in \Omega_{\mathrm{WR}}^{\ast}} I_{\mathrm{WR}}(x,y,\lambda),
\end{equation}
where $N = |\Omega_{\mathrm{WR}}^{\ast}|$. To suppress residual band-to-band jitter without altering the global spectral trend, $\mu_{\mathrm{WR}}(\lambda)$ is smoothed along the spectral axis using a mild 1-D moving-average operator:
\begin{equation}
\tilde{\mu}_{\mathrm{WR}}(\lambda) = \mathrm{Smooth}\!\left[\mu_{\mathrm{WR}}(\lambda)\right].
\end{equation}
Finally, the captured cube is calibrated by band-wise normalization and clipping to the valid reflectance range:
\begin{equation}
I_{\mathrm{cal}}(x,y,\lambda) = \frac{I(x,y,\lambda)}{\tilde{\mu}_{\mathrm{WR}}(\lambda)}, \qquad I_{\mathrm{out}}(x,y,\lambda) = \mathrm{clip}\!\left(I_{\mathrm{cal}}(x,y,\lambda),\,0,\,1\right).
\end{equation}

As shown in \figref{fig:WR_Calibration}, WR normalization reduces residual illumination bias and improves cross-view consistency in pseudo-RGB renderings, producing reconstructions that more closely match the ground-truth appearance.

\begin{algorithm*}[!t]
    \caption{Spectral calibration using a white reference (WR) tarp with automatic spatial mask generation}
    \label{alg:wr_calib}
    \small

    \textbf{Input:}
    \begin{itemize}
      \item Hyperspectral cube \(I(x,y,\lambda)\) with \(L\) spectral bands
      \item Coarse WR region of interest (ROI) \(\Omega_{\mathrm{WR}}\)
      \item Percentile threshold \(p\) for spatial deviation filtering (e.g., \(p = 70\))
      \item 1-D spectral smoothing operator \(\mathrm{Smooth}(\cdot)\)
      \item Clipping bounds \([0,1]\)
    \end{itemize}

    \textbf{Output:}
    \begin{itemize}
      \item Calibrated hyperspectral cube \(I_{\mathrm{out}}(x,y,\lambda)\)
    \end{itemize}

    \textbf{Step 1: WR region extraction.} Extract WR pixels \(I_{\mathrm{WR}}(x,y,\lambda)\) within the coarse ROI \(\Omega_{\mathrm{WR}}\).

    \textbf{Step 2: Spatial deviation analysis.} Compute a per-pixel relative deviation map by averaging over spectral bands:
    \[
    D(x,y) = \frac{1}{L} \sum_{\lambda} \left| \frac{I_{\mathrm{WR}}(x,y,\lambda) - \mu_{\mathrm{WR}}(\lambda)}{\mu_{\mathrm{WR}}(\lambda)} \right|,
    \]
    where \(\mu_{\mathrm{WR}}(\lambda)\) is the preliminary mean WR spectrum.

    \textbf{Step 3: Automatic WR mask generation.} Generate a refined WR mask by retaining pixels whose deviation satisfies
    \[
    D(x,y) \leq \mathrm{Percentile}_p(D),
    \]
    followed by morphological closing, opening, and largest connected component selection to remove edge-affected regions and illumination falloff artifacts.

    \textbf{Step 4: WR mean spectrum computation.} Compute the refined mean WR spectrum over the filtered WR mask:
    \[
    \mu_{\mathrm{WR}}(\lambda) = \frac{1}{N} \sum_{(x,y)\in\Omega_{\mathrm{WR}}^{\ast}} I_{\mathrm{WR}}(x,y,\lambda),
    \]
    where \(N = |\Omega_{\mathrm{WR}}^{\ast}|\).

    \textbf{Step 5: Spectral smoothing.} Smooth the mean WR spectrum along the spectral axis:
    \[
    \tilde{\mu}_{\mathrm{WR}}(\lambda) = \mathrm{Smooth}\!\left[\mu_{\mathrm{WR}}(\lambda)\right].
    \]

    \textbf{Step 6: WR-based normalization.} Normalize each pixel spectrum:
    \[
    I_{\mathrm{cal}}(x,y,\lambda) = \frac{I(x,y,\lambda)}{\tilde{\mu}_{\mathrm{WR}}(\lambda)}.
    \]

    \textbf{Step 7: Final clipping.} Clip calibrated intensities to the valid reflectance range:
    \[
    I_{\mathrm{out}}(x,y,\lambda) = \mathrm{clip}\!\left(I_{\mathrm{cal}}(x,y,\lambda),\,0,\,1\right).
    \]

    \textbf{return} \(I_{\mathrm{out}}(x,y,\lambda)\)
\end{algorithm*}

\subsubsection*{Pose Estimation}
\begin{figure*}[t!]
    \centering
    \includegraphics[width=0.7\linewidth]{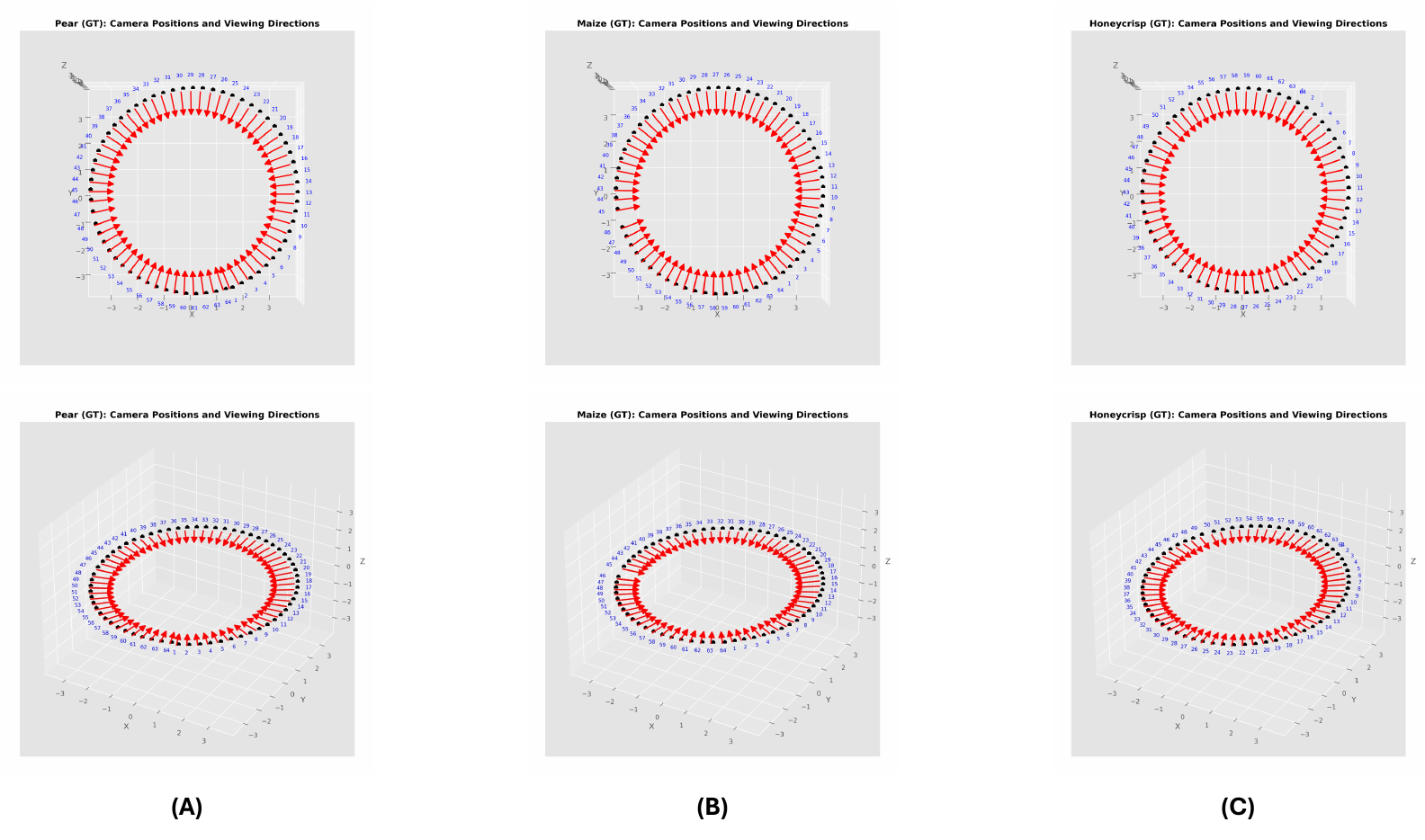}
    \caption{Camera poses and viewing directions estimated by COLMAP for 64 hyperspectral images. Black squares show camera positions, and red arrows indicate viewing directions toward the object center. (A) Pear, (B) Maize, and (C) Apple.}
    \label{fig:colmap_poses}
\end{figure*}

Feature extraction was performed using COLMAP's SIFT feature extractor with GPU acceleration. Sequential matching was then applied to establish correspondences between frames, ensuring temporal consistency. Unlike exhaustive matching, which evaluates all possible image pairs, sequential matching assumes an ordered frame sequence, making it well-suited to the smooth, continuous object rotation in our setup while substantially reducing computational complexity~\citep{feng2025wonderverse}. The resulting camera poses estimated for the hyperspectral image sequence are shown in \figref{fig:colmap_poses}.

The Structure-from-Motion (SfM) pipeline was executed using the COLMAP mapper (v3.12.0), which runs exclusively on the GPU. After evaluating thread counts of 64, 96, and 128, 64 threads were selected as the preferred configuration, as higher counts did not reduce execution time. The sparse point cloud was evaluated based on reprojection error with a target threshold below 1.0~px~\citep{liu2023comparison}, and a bundle adjustment step was applied to jointly refine intrinsic and extrinsic camera parameters.

\subsection{NeRF-Based Hyperspectral PCD Reconstruction}
\label{subsec:nerf}

We present a neural volumetric reconstruction framework for hyperspectral imaging in which a fixed hyperspectral camera observes a rotating object. This setup ensures consistent illumination across views and reduces calibration complexity, making it well-suited for scanning agricultural produce such as fruits, leaves, and other organic specimens in controlled environments. The multiple hyperspectral channels are used jointly to reconstruct a 3D point cloud of the object.

The method builds directly upon the \texttt{nerfacto} pipeline introduced in NeRFStudio~\citep{tancik2023nerfstudio}, extending it to support hyperspectral inputs by modifying the radiance head to predict an $n$-dimensional spectral vector and integrating hyperspectral-specific loss terms. Conceptually, the model follows the HS-NeRF architecture proposed in~\citep{chen2024hyperspectral}, specifically the \textbf{C1, $\sigma_0$} variant from their ablation studies, which directly predicts a multi-channel spectral radiance vector while assuming a scalar, wavelength-invariant transmittance, offering improved stability on noisy data. Whereas HS-NeRF generates novel hyperspectral views, the present work extends it to full 3D reconstruction, producing a dense point cloud with per-point hyperspectral channels.

\subsubsection*{Model Formulation}

Let $\mathbf{x} \in \mathbb{R}^3$ denote a point in 3D space and $\mathbf{d} \in \mathbb{S}^2$ the viewing direction. The network predicts an $n$-channel hyperspectral radiance vector,
\begin{equation}
    \mathbf{c}_\lambda = f_c(\Theta_c(\mathbf{x}, \mathbf{d})) \in \mathbb{R}^n,
\end{equation}
where $n$ is the number of hyperspectral channels, and a scalar transmittance value invariant across wavelength,
\begin{equation}
    \bar{\sigma} = f_\sigma(\Theta_\sigma(\mathbf{x})) \in \mathbb{R}.
\end{equation}
This simplified formulation retains the benefits of a volumetric rendering framework while reducing the cost and instability often associated with learning continuous spectral functions.

\subsubsection*{Training Protocol}

\begin{figure*}[t!]
    \centering
    \includegraphics[width=0.99\linewidth]{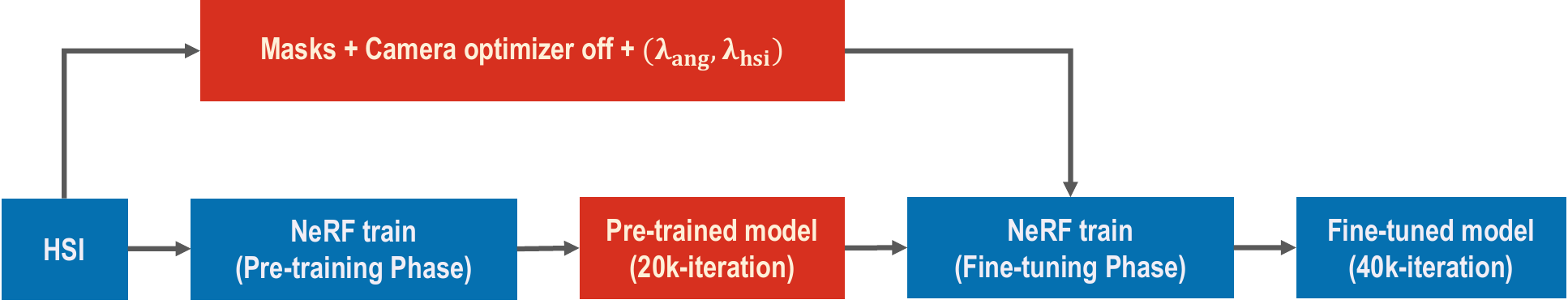}
    \caption{HSI-SC-NeRF two-stage training protocol. The model is first pre-trained on full-frame HSI, then fine-tuned with foreground masks and the camera optimizer disabled using the selected loss-weight setting $(\lambda_{\mathrm{ang}}, \lambda_{\mathrm{hsi}})$. Iteration counts are shown in the diagram.}
    \label{fig:ablation_study_pipeline}
\end{figure*}

\figref{fig:ablation_study_pipeline} summarizes the two-stage training procedure. The hyperspectral NeRF is first pre-trained on full-frame inputs to obtain a stable initialization for geometry and coarse spectral alignment, using all pixels in each view so that the model learns a consistent volumetric representation of the scene's global radiance distribution. The pre-trained checkpoint is then fine-tuned using foreground ROI masks to restrict supervision to object pixels, reducing background-driven gradients and concentrating optimization on radiometrically meaningful regions. Camera optimization is disabled during fine-tuning to decouple radiometric refinement from pose updates and isolate the effect of hyperspectral supervision. Across both stages, a weighted combination of spectral angular loss and magnitude loss is applied, controlled by $(\lambda_{\mathrm{ang}}, \lambda_{\mathrm{hsi}})$; the final setting is selected via the loss-weight sensitivity analysis in \secref{subsec:ablation_loss_weights_maize}.

\subsubsection*{Loss Function}

The hyperspectral NeRF is trained using a composite objective that combines Nerfacto's default volumetric rendering losses from Nerfstudio with hyperspectral-specific spectral supervision:
\begin{align}
    \mathcal{L} =& \lambda_{\text{hsi}} \, \mathcal{L}_{\text{hsi}} + \lambda_{\text{ang}} \, \mathcal{L}_{\text{ang}} + \lambda_{\text{prop}} \, \mathcal{L}_{\text{prop}}  \\ \nonumber
    + &\lambda_{\text{dist}} \, \mathcal{L}_{\text{dist}} + \lambda_{\text{ori}} \, \mathcal{L}_{\text{ori}} + \lambda_{\text{pn}} \, \mathcal{L}_{\text{pn}}.
\end{align}

\textbf{Hyperspectral Reconstruction Loss} ($\mathcal{L}_{\text{hsi}}$) is the primary data fidelity term, enforcing accurate recovery of per-pixel spectral radiance across all $n$ wavelength channels via mean squared error over spectral bands:
\begin{equation}
\mathcal{L}_{\text{hsi}} = \frac{1}{|\Omega|}\sum_{p\in\Omega}\sum_{i=1}^{n} \left( \hat{c}^{(p)}_{\lambda_i} - c^{(p)}_{\lambda_i} \right)^2,
\end{equation}
where $p$ indexes pixels and $\Omega$ is the set of valid pixels, optionally selected by a foreground mask.

\textbf{Angular Spectral Loss} ($\mathcal{L}_{\text{ang}}$) encourages agreement in spectral shape independent of overall magnitude by penalizing the angular discrepancy between predicted and ground-truth spectral vectors via cosine similarity. For each valid pixel $p$, let $\hat{\mathbf{s}}(p), \mathbf{s}(p)\in\mathbb{R}^{204}$ denote the predicted and target spectra. The cosine similarity is
\begin{equation}
s(p) = \frac{\left\langle \hat{\mathbf{s}}(p), \mathbf{s}(p) \right\rangle}{\left\|\hat{\mathbf{s}}(p)\right\|_2\,\left\|\mathbf{s}(p)\right\|_2+\epsilon},
\end{equation}
and the loss is defined as
\begin{equation}
\mathcal{L}_{\text{ang}} = \frac{1}{|\Omega|}\sum_{p\in\Omega}\left(1-s(p)\right),
\label{eq:ang_cosine_loss}
\end{equation}
where $\epsilon$ is a small constant for numerical stability.

\textbf{Proposal Loss} ($\mathcal{L}_{\text{prop}}$), following MipNeRF-360~\citep{barron2022mip}, provides coarse-to-fine proposal supervision to stabilize ray sampling and accelerate convergence, which is particularly beneficial for high-channel data where sampling inefficiencies compound during training.

\textbf{Distortion Regularization} ($\mathcal{L}_{\text{dist}}$), following NeRF++ and Inst\-ant-NGP~\citep{muller2022instant}, prevents sample collapse along viewing rays by penalizing excessive contraction of sample intervals, and additionally discourages spectral aliasing due to tightly clustered points.

\textbf{Orientation Loss} ($\mathcal{L}_{\text{ori}}$) encourages surface normals inferred from volumetric gradients to align with known viewing directions, improving rendering quality and view-consistency on smooth surfaces:
\begin{equation}
\mathcal{L}_{\text{ori}} = \sum_{i} \left(1 - \langle \hat{n}_i, v_i \rangle^2 \right),
\end{equation}
where $\hat{n}_i$ is the predicted normal and $v_i$ is the viewing vector.

\textbf{Predicted-Normal Loss} ($\mathcal{L}_{\text{pn}}$) supervises a normal prediction head by matching predicted normals to computed normals derived from volumetric density gradients, following the Nerfstudio Nerfacto pipeline. For each ray, the predicted normal $\tilde{\mathbf{n}}_{j}$ at sample $j$ is compared to the computed normal $\hat{\mathbf{n}}_{j}$, weighted by the volumetric rendering weight $w_j$:
\begin{equation}
\mathcal{L}_{\text{pn}} = \mathbb{E}_{\text{rays}} \left[ \sum_{j} w_j \left( 1 - \hat{\mathbf{n}}_{j}^{\top}\tilde{\mathbf{n}}_{j} \right) \right],
\end{equation}
where higher weights correspond to samples that more strongly explain the pixel color. This loss improves surface regularity and view-consistent rendering along high-opacity regions.

In all experiments, Nerfacto's default loss multipliers from Nerfstudio are retained: $\lambda_{\text{prop}}{=}1.0$, $\lambda_{\text{dist}}{=}0.002$, $\lambda_{\text{ori}}{=}10^{-4}$, and $\lambda_{\text{pn}}{=}10^{-3}$. Only the weights of the two hyperspectral-specific terms, $\lambda_{\text{hsi}}$ and $\lambda_{\text{ang}}$, are tuned, using the cosine variant of $\mathcal{L}_{\text{ang}}$ throughout (\texttt{angu\-lar\_loss\_type=cosine}).

\subsection{Evaluation Metrics}
\label{sec:metrics}

\subsubsection*{Spatial Evaluation Metrics}

The quality of reconstructed point clouds is quantitatively assessed using precision and recall computed after ground-truth alignment~\citep{arshad2024evaluating}. Let $\mathcal{P}_{\mathrm{sc}}$ denote the set of stationary-camera NeRF reconstructed points and $\mathcal{P}_{\mathrm{gt}}$ the set of ground-truth points obtained from a standard moving-camera NeRF reconstruction. Precision is the fraction of reconstructed points within a threshold distance $\epsilon$ of any ground-truth point,
\[
\text{Precision}(\epsilon) = \frac{\left|\left\{ x \in \mathcal{P}_{\mathrm{sc}} \mid \min_{y \in \mathcal{P}_{\mathrm{gt}}} \|x - y\| \leq \epsilon \right\}\right|}{\left|\mathcal{P}_{\mathrm{sc}}\right|},
\]
and recall is the fraction of ground-truth points that have a corresponding reconstructed point within the same threshold,
\[
\text{Recall}(\epsilon) = \frac{\left|\left\{ y \in \mathcal{P}_{\mathrm{gt}} \mid \min_{x \in \mathcal{P}_{\mathrm{sc}}} \|y - x\| \leq \epsilon \right\}\right|}{\left|\mathcal{P}_{\mathrm{gt}}\right|}.
\]
Together, these metrics capture both the inclusion of geometrically relevant surface detail and the exclusion of spurious points.

\subsubsection*{Spectral Evaluation Metrics}

Reconstruction quality on held-out views is evaluated using hyperspectral metrics computed directly on the full spectral cube (204 bands). Let $\hat{\mathbf{s}}(p)\in\mathbb{R}^{204}$ and $\mathbf{s}(p)\in\mathbb{R}^{204}$ denote the predicted and ground-truth spectral vectors at pixel $p$, and let $\Omega$ be the set of evaluated pixels, optionally restricted by a foreground mask. All metrics are reported as mean $\pm$ standard deviation across held-out views. Spectral fidelity is quantified via spectral RMSE and the spectral angle mapper (SAM), which serve as primary metrics. Hyperspectral PSNR and SSIM are additionally reported as complementary spatial consistency indicators, computed per band and averaged across wavelengths by extending Nerfstudio's default evaluation pipeline to the full 204-band hyperspectral renderings~\citep{tancik2023nerfstudio}.

\textbf{SAM (Spectral Angle Mapper)}~\citep{kruse1993sam} measures the angular discrepancy between predicted and ground-truth spectra and is scale-invariant with respect to multiplicative intensity changes:
\begin{equation}
\mathrm{SAM} = \frac{1}{|\Omega|}\sum_{p\in\Omega} \arccos\!\left(\frac{\langle \hat{\mathbf{s}}(p), \mathbf{s}(p)\rangle}{\|\hat{\mathbf{s}}(p)\|_2\,\|\mathbf{s}(p)\|_2+\delta}\right),
\end{equation}
where $\delta$ is a small constant for numerical stability.

\textbf{Spectral RMSE} quantifies cube-level radiometric error. The spectral mean squared error is defined as
\begin{equation}
\mathrm{MSE}_{\text{spectral}} = \frac{1}{|\Omega|}\sum_{p\in\Omega}\frac{1}{204}\left\|\hat{\mathbf{s}}(p)-\mathbf{s}(p)\right\|_2^2,
\end{equation}
and spectral RMSE is computed as $\mathrm{RMSE}_{\text{spectral}} = \sqrt{\mathrm{MSE}_{\text{spectral}}}$.

\textbf{HSI-SSIM}~\citep{wang2004ssim} is computed per band and averaged across wavelengths:
\begin{equation}
\mathrm{SSIM}_{\text{HSI}} = \frac{1}{204}\sum_{b=1}^{204}\mathrm{SSIM}\!\left(\hat{S}_b, S_b\right),
\end{equation}
where the standard single-band SSIM is
\begin{equation}
\mathrm{SSIM}(\hat{S}_b,S_b) = \frac{(2\mu_{\hat{S}_b}\mu_{S_b}+C_1)(2\sigma_{\hat{S}_b S_b}+C_2)}{(\mu_{\hat{S}_b}^2+\mu_{S_b}^2+C_1)(\sigma_{\hat{S}_b}^2+\sigma_{S_b}^2+C_2)},
\end{equation}
with $\mu_{\cdot}$, $\sigma_{\cdot}^2$, and $\sigma_{\hat{S}_b S_b}$ computed over pixels in $\Omega$, and $C_1, C_2$ stability constants as in \citet{wang2004ssim}.

\textbf{HSI-PSNR}~\citep{huynh2008psnr} is computed per band from the per-band MSE and averaged across wavelengths:
\begin{align}
\mathrm{MSE}_b = \frac{1}{|\Omega|}\sum_{p\in\Omega}\left(\hat{S}_b(p)-S_b(p)\right)^2,\\ \mathrm{PSNR}_{\text{HSI}} = \frac{1}{204}\sum_{b=1}^{204}10\log_{10}\!\left(\frac{I_{\max}^2}{\mathrm{MSE}_b+\delta}\right),
\end{align}
with $I_{\max}=1$ after normalization and $\delta$ for numerical stability.

\section{Results and Discussion}
\label{sec:Results}

The HSI-SC-NeRF framework is evaluated under criteria relevant to agricultural applications, encompassing spatial reconstruction accuracy, wavelength-resolved spectral fidelity, and sensitivity to loss-weight configuration. The model predicts $n$-channel radiance vectors with a wavelength-invariant density field, following the C1, $\sigma_0$ configuration of~\citet{chen2024hyperspectral}, prioritizing robustness, computational efficiency, and noise tolerance for practical deployment.

\subsection{Example Applications}
\label{subsec:applications}

The reconstructed 3D hyperspectral point clouds are used to support spatially resolved comparisons of spectral signatures and band-specific volumetric comparisons under a consistent processing pipe\-line (\figref{fig:Workflow}). For each crop, meaningful spectral bands are selected based on prior studies linking specific wavelengths to key biochemical and physical indicators. These bands are not used during training but serve exclusively for post-reconstruction interpretation of reflectance patterns.

\begin{itemize}[left=0pt]
    \item \textbf{Apple}: 551~\unit{nm} (green band, sensitive to chlorophyll and browning pigments in the peel); 708~\unit{nm} (red-edge, sensitive to pigment, structure, and water status, and commonly selected for fruit maturity and stress assessment); 801~\unit{nm} (NIR band in the 800--820~\unit{nm} region recommended for apple bruise and defect detection)~\citep{baek2019selection, baek2022determination, fsian2025apple}.

    \item \textbf{Maize Kernel}: 490~\unit{nm} (pigment variation among cultivars); 580~\unit{nm} (carotenoid content and maturity); 850~\unit{nm} (dryness and starch accumulation)~\citep{alimohammadi2022hyperspectral, thenkabail2000hyperspectral}.

    \item \textbf{Pear}: 550~\unit{nm} (green band, sensitive to peel color and chlorophyll content); 710~\unit{nm} (red-edge, linked to pigment changes, maturity, and internal quality); 980~\unit{nm} (NIR water-related band associated with internal water status, firmness, and bruise or disease detection)~\citep{lee2014hyperspectral, cruz2021nondestructive}.
\end{itemize}

These post-reconstruction spectral signatures provide biologically relevant cues for interpreting crop condition and quality, enabling targeted, wavelength-aware evaluation of NeRF-generated reconstructions in an agricultural context.

\subsection{Spatial Validation}
\label{subsec:spatial_val}

\begin{figure*}[t!]
    \centering
    \includegraphics[width=0.9\linewidth]{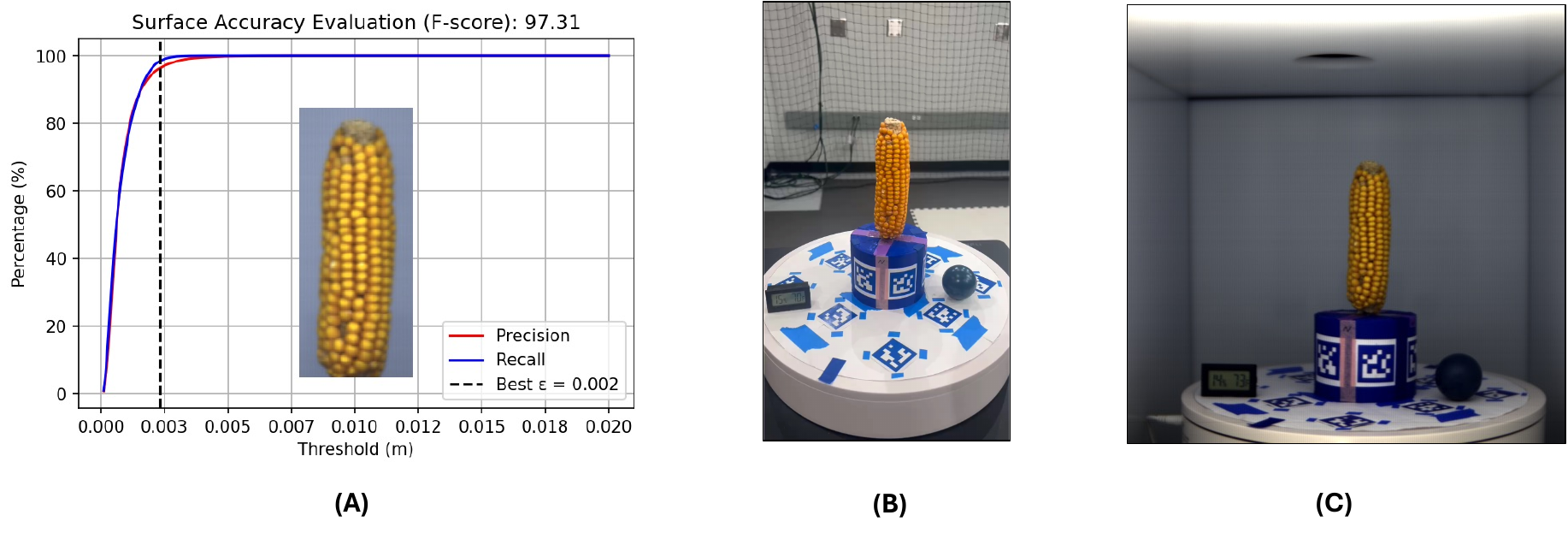}
    \caption{Spatial validation of maize ear reconstruction. (A) Precision--recall curves as functions of the distance threshold $\epsilon$, with the optimal threshold ($\epsilon = 0.002$~m) indicated by the black dashed line, yielding an F-score of 97.31. Point-to-surface distances were computed after ICP registration in CloudCompare, resulting in a final RMS error of $0.001$~m over approximately 50{,}000 points. (B) Reference PCD obtained using a rotating RGB camera (iPhone~13~Pro, $2532 \times 1170$). (C) Predicted PCD generated from hyperspectral imagery captured inside the Teflon imaging chamber using a stationary SPECIM camera ($512 \times 512$).}
    \label{fig:precision-recall}
\end{figure*}

Spatial reconstruction accuracy was evaluated using a precision--recall analysis based on point-to-surface distances between the reconstructed maize ear and the reference model. After rigid alignment via iterative closest point (ICP) registration in CloudCompare, precision and recall were computed as functions of a distance threshold $\epsilon$, defining the maximum allowable deviation for a predicted surface point to be considered a correct correspondence. As shown in \figref{fig:precision-recall}, the optimal threshold was identified at $\epsilon = 0.002$~m (2~mm), at which the reconstruction achieved an F-score of 97.31, indicating strong geometric agreement with the reference. This threshold is conservative relative to the typical kernel thickness of maize ear structures ($\approx$5~mm). ICP registration converged with a final RMS point-to-point error of $0.001$~m over approximately 50{,}000 sampled points. The reference model was reconstructed using a rotating RGB camera (iPhone~13~Pro, $2532 \times 1170$), while the evaluated reconstruction was generated from hyperspectral imagery captured with a stationary SPECIM camera ($512 \times 512$).

\subsection{Spectral Validation}
\label{subsec:spectral_validation}

This section evaluates whether the proposed hyperspectral NeRF preserves wavelength-resolved spectral behavior under radiometrically calibrated inputs. All results correspond to the final training configuration $(\lambda_{\mathrm{ang}}, \lambda_{\mathrm{hsi}})=(0.25, 0.75)$ selected by the ablation study (\secref{subsec:ablation_loss_weights_maize}), with each dataset split into 90\% training views and 10\% held-out views. All reported metrics are computed directly on the full 204-band hyperspectral renderings and their corresponding held-out ground truth.

\begin{figure*}[t!]
    \centering
    \includegraphics[width=0.9\linewidth]{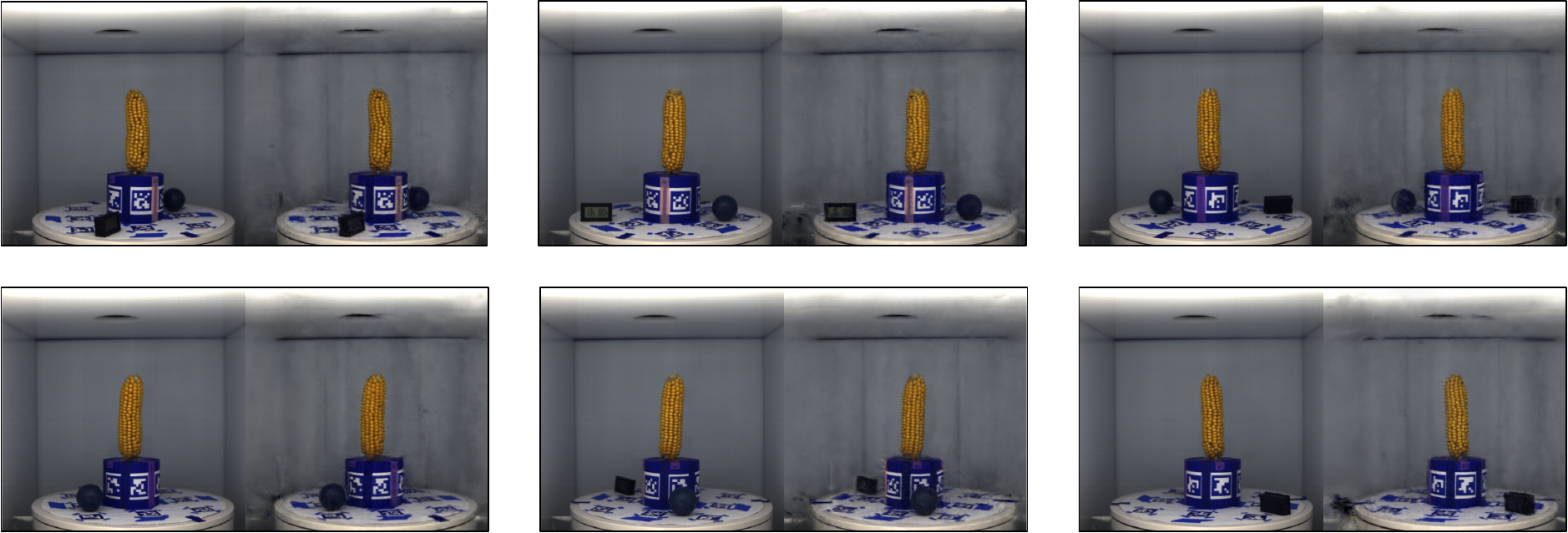}
    \caption{Evaluation images. Six representative frames (10\% of the 64 total frames) are shown. In each maize ear, the left image shows the ground-truth reference and the right image shows the prediction.}
    \label{fig:maize_eval}
\end{figure*}

\begin{table*}[t]
\centering
\caption{HSI evaluation metrics with the loss-weight setting $(\lambda_{\mathrm{ang}}, \lambda_{\mathrm{hsi}})=(0.25,0.75)$ on the 204-band hyperspectral datasets. SAM, RMSE, PSNR, and SSIM are computed directly over the full 204-band hyperspectral renderings. Values are reported as mean $\pm$ SD over held-out views.}
\label{tab:HSI_eval_metrics_final}
\renewcommand{\arraystretch}{1.5}
\setlength{\tabcolsep}{14pt}
\begin{tabular}{lccc}
\toprule
\multicolumn{1}{c}{\large HSI Eval Metrics} & \large Maize & \large Pear & \large Apple \\
\midrule
SAM [rad] $\downarrow$ & $0.02921 \pm 0.00457$ & $0.04577 \pm 0.01281$ & $0.06420 \pm 0.03807$ \\
RMSE $\downarrow$      & $0.03051 \pm 0.01141$ & $0.03242 \pm 0.00754$ & $0.04383 \pm 0.02415$ \\
SSIM $\uparrow$      & $0.88823 \pm 0.05618$ & $0.83259 \pm 0.03064$ & $0.74229 \pm 0.11584$ \\
PSNR [dB] $\uparrow$ & $29.13709 \pm 2.55173$ & $27.47112 \pm 3.20361$ & $24.49213 \pm 5.85267$ \\
\midrule
Number of evaluation rays & \multicolumn{3}{c}{$262{,}144$} \\
\bottomrule
\end{tabular}
\end{table*}

\tabref{tab:HSI_eval_metrics_final} summarizes quantitative performance on held-out views for all three crops under the loss-weight setting $(\lambda_{\mathrm{ang}}, \lambda_{\mathrm{hsi}})= (0.25,$ $0.75)$. Unlike conventional NeRF evaluation reporting PSNR, SSIM, or LPIPS on RGB images, hyperspectral PSNR and SSIM are computed directly over the full 204-band outputs (band-wise then averaged), together with SAM and spectral RMSE, to assess wavelength-resolved fidelity.

SAM remains well below the default ENVI classification threshold of 0.1~rad~\citep{envi2014sam} across all datasets~\citep{kruse1993sam}, and spectral RMSE remains low, indicating low spectral-shape discrepancy and low absolute per-band error across wavelengths. Hyperspectral PSNR and SSIM are consistent with these spectral results and fall within or near the range commonly reported for high-quality NeRF reconstructions on standard benchmarks~\citep{mildenhall2021nerf, xie2023s3im}, though they are treated as supportive indicators rather than primary criteria, as they can be sensitive to evaluation details and do not fully capture wavelength-resolved fidelity~\citep{tancik2023nerfstudio}. Note that PSNR/SSIM values computed on 204-band hyperspectral renderings are not directly comparable to RGB-only benchmark reports.

The metrics exhibit a consistent dataset difficulty ordering of Maize $\rightarrow$ Pear $\rightarrow$ Apple across all evaluation criteria. This trend is attributable to crop-specific differences in geometry and surface appearance, including shape complexity, texture, and wavelength-dependent reflectance variation such as localized bruising patterns and stronger view-dependent effects, all of which increase reconstruction ambiguity and reduce spatial and spectral agreement.

Qualitative comparisons in \figref{fig:maize_eval} further support these trends, showing that predicted pseudo-RGB renderings closely match held-out ground-truth images across representative viewpoints, with remaining discrepancies primarily occurring around fine structures and view-dependent highlights.


\subsection{Ablation on Loss-Weight Sensitivity}
\label{subsec:ablation_loss_weights_maize}

To study how hyperspectral 3D reconstruction responds to supervision balance, a controlled loss-weight ablation was performed on the maize ear dataset over five configurations: $(\lambda_{\text{ang}}, \lambda_{\text{hsi}}) \in \{(0,1), (0.25,0.75), (0.5,0.5), (0.75,0.25), (1,0)\}$. As shown in \figref{fig:ablation_study_pipeline}, all models were first pre-trained on full frames for 20k iterations to obtain a stable initialization for geometry and coarse spectral alignment, then fine-tuned for an additional 20k iterations using foreground masks to concentrate gradients on object pixels. Camera optimization was disabled during fine-tuning to decouple radiometric refinement from pose updates, enabling a cleaner interpretation of how loss-weighting affects hyperspectral fidelity.

\begin{table*}[t]
    \centering
    \caption{Pre-trained vs.\ fine-tuned results on the maize dataset across loss weights $(\lambda_{\mathrm{ang}}, \lambda_{\mathrm{hsi}})$, reported as mean $\pm$ SD over held-out views. Metrics are grouped into spectral fidelity (SAM, RMSE; lower is better) and spatial/image-quality proxies (SSIM, PSNR; higher is better). Best fine-tuned values are highlighted in bold.}
    \label{tab:maize_pre_ft_meanpmstd}
    \setlength{\extrarowheight}{4pt}
    \resizebox{\textwidth}{!}{
        \begin{tabular}{l|cc|cc|cc|cc}
        \hline
        \textbf{(Angular, HSI)}
        & \multicolumn{2}{c|}{\textbf{SAM [0--1]} $\downarrow$}
        & \multicolumn{2}{c|}{\textbf{RMSE [0--1]} $\downarrow$}
        & \multicolumn{2}{c|}{\textbf{SSIM [0--1]} $\uparrow$}
        & \multicolumn{2}{c}{\textbf{PSNR [dB]} $\uparrow$} \\
        \cline{2-9}
        & Pre-train & Fine-tune & Pre-train & Fine-tune & Pre-train & Fine-tune & Pre-train & Fine-tune \\
        \hline
        (0,1) HSI-only
        & $0.0622 \pm 0.0113$ & $0.0422 \pm 0.0192$
        & $0.0731 \pm 0.0134$ & $0.0360 \pm 0.0141$
        & $0.3905 \pm 0.1377$ & $0.8564 \pm 0.0718$
        & $22.11 \pm 1.63$ & $27.27 \pm 4.43$ \\
        \textbf{(0.25,0.75)}
        & $0.0588 \pm 0.0091$ & $\mathbf{0.0292 \pm 0.0046}$
        & $0.0715 \pm 0.0098$ & $\mathbf{0.0305 \pm 0.0114}$
        & $0.4356 \pm 0.1161$ & $\mathbf{0.8882 \pm 0.0562}$
        & $22.17 \pm 1.34$ & $\mathbf{29.14 \pm 2.55}$ \\
        (0.5,0.5)
        & $0.0808 \pm 0.0130$ & $0.0523 \pm 0.0204$
        & $0.0922 \pm 0.0085$ & $0.0468 \pm 0.0162$
        & $0.1758 \pm 0.0972$ & $0.7896 \pm 0.0978$
        & $19.81 \pm 1.11$ & $24.45 \pm 3.59$ \\
        (0.75,0.25)
        & $0.0498 \pm 0.0087$ & $0.0428 \pm 0.0169$
        & $0.0616 \pm 0.0141$ & $0.0421 \pm 0.0138$
        & $0.5212 \pm 0.1676$ & $0.8534 \pm 0.0802$
        & $23.78 \pm 1.91$ & $25.89 \pm 3.94$ \\
        (1,0) Angular-only
        & $0.0707 \pm 0.0249$ & $0.0381 \pm 0.0162$
        & $0.0893 \pm 0.0120$ & $0.0732 \pm 0.0096$
        & $0.2811 \pm 0.1276$ & $0.5984 \pm 0.0342$
        & $19.97 \pm 1.51$ & $21.65 \pm 1.69$ \\
        \hline
        \end{tabular}
    }
\end{table*}

\paragraph{Evaluation Metric Categorization: Spectral Fidelity vs.\ Spatial Proxies}
Both pre-trained and fine-tuned models are evaluated using four metrics: SAM, spectral RMSE, SSIM, and PSNR (\tabref{tab:maize_pre_ft_meanpmstd}), grouped by their physical relevance to hyperspectral analysis. SAM and spectral RMSE are treated as primary metrics, as they directly quantify wavelength-resolved agreement in spectral shape and magnitude respectively. SSIM and PSNR are reported as secondary spatial proxies. Although standard in NeRF evaluation, PSNR and SSIM do not inherently guarantee wavelength-resolved spectral fidelity: because NeRF-style volume rendering integrates transmittance-weighted radiance samples along a ray~\citep{mildenhall2021nerf} and is commonly optimized with L2-type objectives, it can favor solutions that minimize average pixel-wise error without corresponding gains in spectral correctness~\citep{amir2021understanding}. SAM and spectral RMSE are therefore emphasized as the primary criteria throughout.

\paragraph{Pre-training Phase}
In the full-frame pre-training regime, the angular-dominant hybrid $(0.75, 0.25)$ yields the best spectral fidelity, while the balanced setting $(0.5, 0.5)$ performs worst. This ordering is consistent with the known behavior of SAM: as a measure of angle between spectral vectors, it primarily captures spectral shape agreement and is relatively insensitive to illumination scaling on calibrated reflectance~\citep{kruse1993sam}. Background pixels and view-dependent intensity variation tend to dominate early gradients in full-frame training; emphasizing an angular constraint therefore stabilizes wavelength-wise alignment of the object's spectral signature, while the magnitude term anchors the intensity scale. Pure angular supervision $(1, 0)$ does not outperform the best hybrid, suggesting that an explicit intensity anchor remains beneficial to limit scale ambiguity during early optimization.

\paragraph{Fine-tuning Phase}
After fine-tuning, the optimal configuration shifts to the magnitude-dominant hybrid $(0.25, 0.75)$, exhibiting a clear rank shift relative to pre-training. Once supervision is concentrated on object pixels and pose refinement is disabled, the remaining error becomes primarily radiometric; increasing $\lambda_{\text{hsi}}$ more effectively reduces per-wavelength intensity discrepancies, while retaining a small angular component preserves spectral-shape structure. The angular-only setting $(1, 0)$ shows comparatively strong shape agreement but weaker magnitude accuracy, confirming that scale-invariant objectives alone do not guarantee radiometric correctness. This is consistent with prior work showing that joint pose optimization can absorb photometric error during NeRF training, and that disabling camera optimization in the fine-tuning stage helps isolate radiometric refinement~\citep{lin2021barf, takeda2023photometricba}.

Overall, \tabref{tab:maize_pre_ft_meanpmstd} confirms that hybrid supervision is more reliable than single-loss baselines, as it jointly constrains spectral shape and magnitude. The optimal weighting is stage-dependent: full-frame pre-training benefits from angular-regularized hybrids, whereas masked fine-tuning with camera optimization disabled favors magnitude-dominant refinement. The configuration $(0.25, 0.75)$ is selected as the final setting for the maize ear dataset, prioritizing wavelength-resolved hyperspectral fidelity.

\subsection{Applications: Band-Dependent Visualization and Spectral Separability}
\label{subsec:band_viz}

To illustrate how hyperspectral reconstructions support application-driven interpretation, a band-dependent visualization analysis was conducted on three produce samples (apple, maize ear, and pear). The central hypothesis is that selecting wavelength triplets targeting biologically and chemically driven reflectance differences can increase spectral separability between tissue states and, in some cases, propagate to downstream geometry-based traits derived from meshing. For each sample, a baseline pseudo-RGB rendering is compared with an interpretation-driven 3-channel composite; ROI-level spectra are visualized for healthy versus affected regions; and Poisson mesh surface area and volume are reported to assess sensitivity of trait estimates to band selection.

\begin{figure*}[t!]
    \centering
    \includegraphics[width=0.99\textwidth]{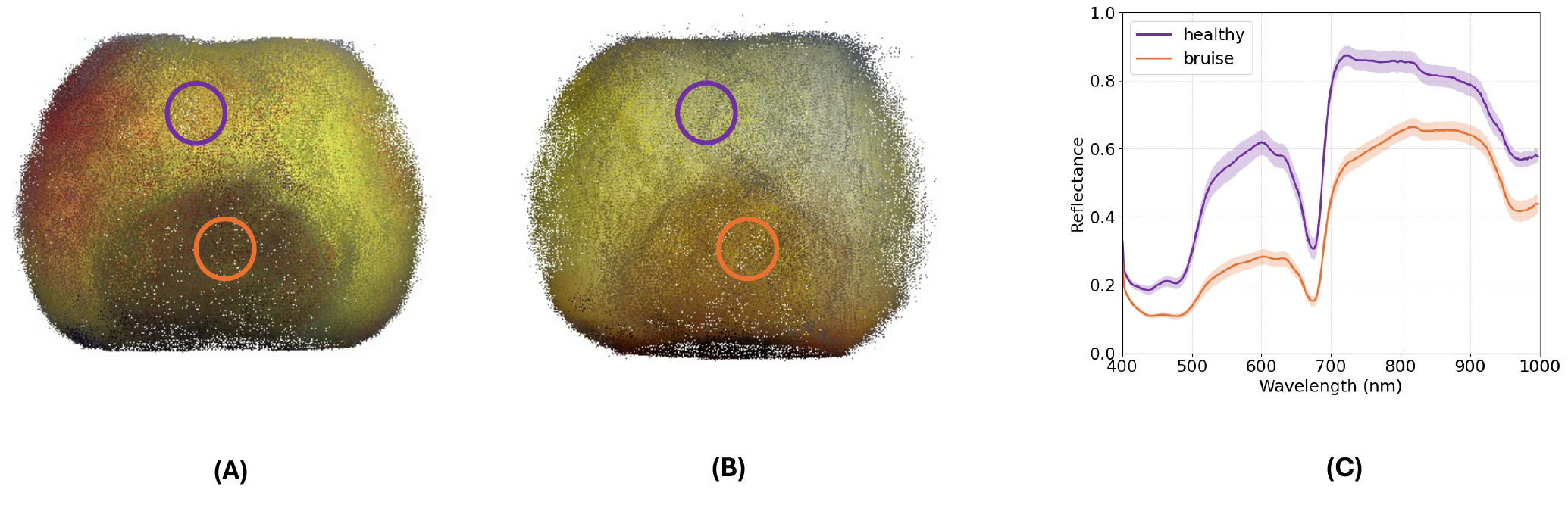}
    \caption{Apple visualization and spectra. (A) Pseudo-RGB (598, 548, 449~nm). (B) Bruise-contrast composite (801, 708, 551~nm). (C) Mean reflectance spectra from the marked healthy and bruised regions.}
    \label{fig:app_apple_bruise}
\end{figure*}

\paragraph{Apple}
A standard pseudo-RGB composite is compared with a bruise-oriented triplet designed to emphasize bruise-sensitive wavelength regions. As shown in \figrefA{fig:app_apple_bruise}{A--B}, the bruise-oriented composite (801, 708, 551~nm) increases contrast of the affected region relative to pseudo-RGB. The ROI spectra (\figrefA{fig:app_apple_bruise}{C}) indicate consistent reflectance separability between healthy and bruised tissue in the visible range (550--650~nm) and in the red-edge/NIR region (708--801~nm), consistent with wavelength-dependent responses to tissue condition. Poisson surface reconstructions (octree depth 8) generated from each composite show that the bruise-oriented composite yields a smaller surface area and larger volume compared to pseudo-RGB (\tabref{tab:mesh_area_volume_all_bands}), suggesting that band choices altering material-dependent contrast can also affect mesh-derived trait estimates.

\begin{figure*}[t!]
    \centering
    \includegraphics[width=0.99\textwidth]{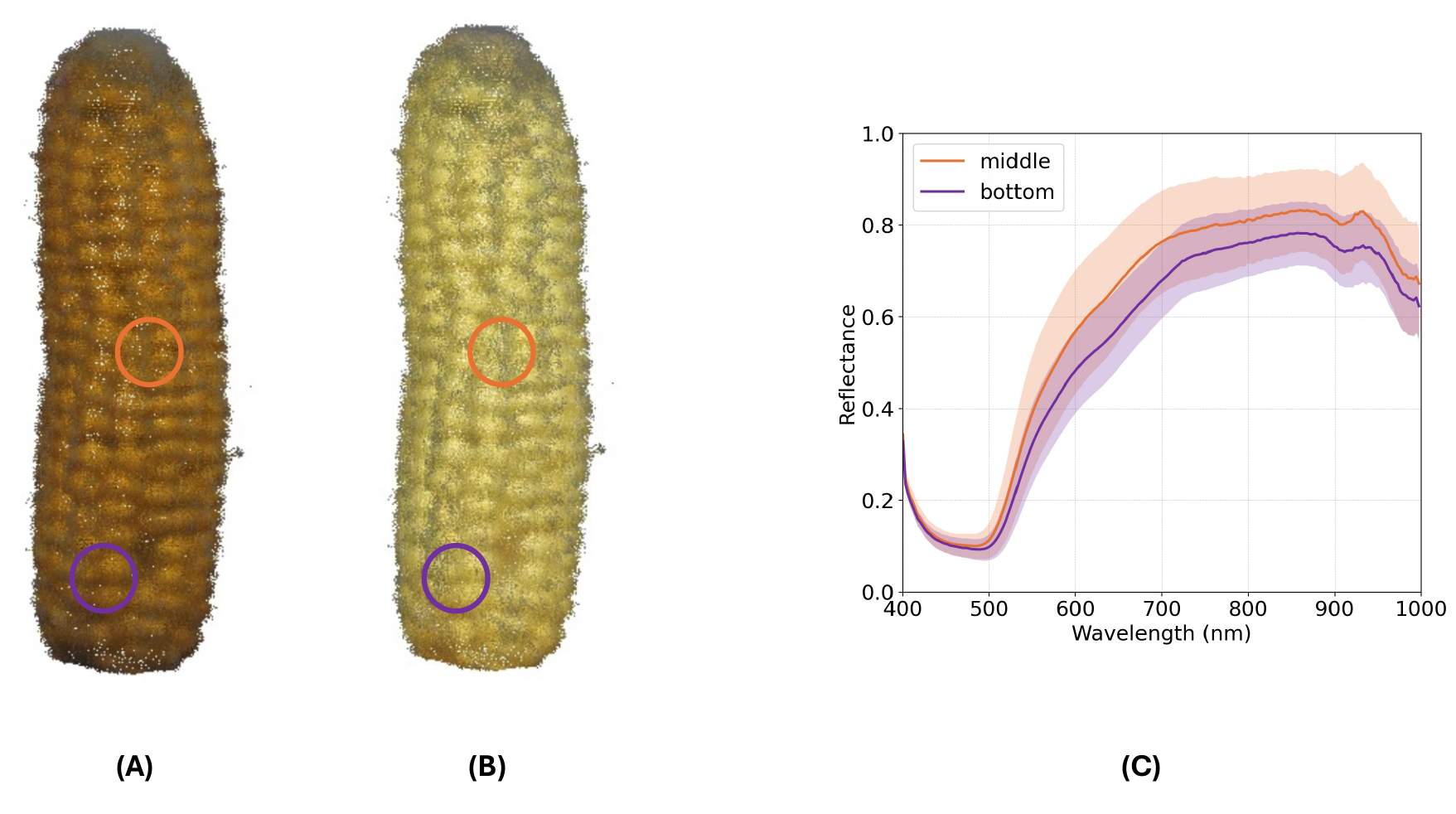}
    \caption{Maize reconstructions under different band triplets. (A) Visible composite [650, 540, 470]~\unit{nm}. (B) NIR-inclusive maize triplet [850, 580, 490]~\unit{nm}. (C) Spectral reflectance curves extracted from the highlighted surface regions in (A) and (B).}
    \label{fig:app_maize}
\end{figure*}

\paragraph{Maize}
The maize ear is visualized using two 3-channel composites: a visible-band triplet and an NIR-inclusive triplet selected to highlight kernel-level contrast, rendered at [650, 540, 470]~\unit{nm} and [850, 580, 490]~\unit{nm} respectively (\figrefA{fig:app_maize}{A--B}). The ROI spectra (\figrefA{fig:app_maize}{C}) show localized reflectance variation across anatomical features spanning both the visible and NIR wavelengths. Mesh-derived traits computed from Poisson reconstructions (octree depth 8) are identical across the two composites (\tabref{tab:mesh_area_volume_all_bands}), indicating that while band selection changes visual appearance and spectral contrast, the global mesh geometry remains stable under the current meshing configuration.

\begin{figure*}[t!]
    \centering
    \includegraphics[width=0.99\linewidth]{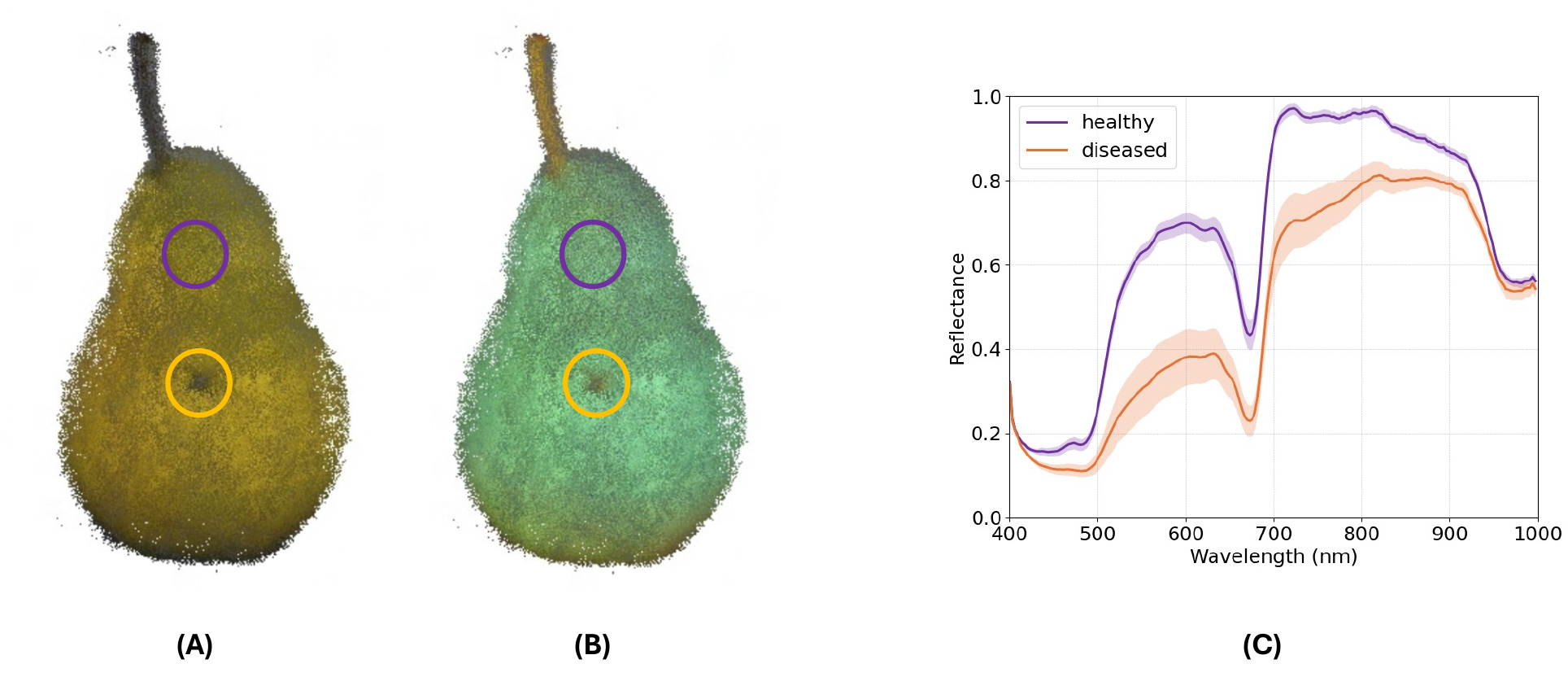}
    \caption{Pear visualization and spectra. (A) Pseudo-RGB composite. (B) Disease-oriented composite (980, 710, 550~nm). (C) Mean reflectance spectra from the marked healthy and diseased regions.}
    \label{fig:app_pear}
\end{figure*}

\paragraph{Pear}
A pseudo-RGB composite is compared with an interpre\-tation-driven triplet selected to emphasize peel-color, red-edge, and water-related responses. The disease-oriented composite (980, 710, 550~nm) increases visibility of quality differences relative to pseudo-RGB (\figrefA{fig:app_pear}{A--B}). The ROI spectra (\figref{fig:app_pear}{C}) show separability between healthy and diseased tissue near 550~nm, 710~nm, and 980~nm, consistent with wavelength-dependent responses to tissue composition and water status. Surface area and volume estimates from Poisson meshes (octree depth 8) remain nearly unchanged across composites (\tabref{tab:mesh_area_volume_all_bands}), suggesting that band selection can improve spectral separability without perturbing global mesh traits for this sample.

Overall, interpretation-driven band selection enhances spectral separability between biologically distinct tissues and, in some cases, produces measurable differences in mesh-derived geometric traits. The apple case demonstrates that band choices aligned with chemically and structurally driven reflectance changes can propagate to downstream surface area and volume estimates, whereas the maize and pear examples show comparatively stable global geometry under the same meshing configuration. These results motivate reporting both ROI-level spectra and geometry-based traits when hyperspectral reconstructions are used for phenotyping or postharvest assessment.

\begin{table}[t!]
    \centering
    \caption{Surface area and volume estimates for Poisson meshes (octree depth 8) reconstructed from different 3-channel composites (Figs.~\ref{fig:app_apple_bruise}--\ref{fig:app_pear}A--B). Area is reported in m$^2$ and volume in liters (L).}
    \label{tab:mesh_area_volume_all_bands}
    \small
    \begin{tabular}{llrr}
    \toprule
    \textbf{Crop} & \textbf{Band composite [nm]} & \textbf{Area (m$^2$)} & \textbf{Volume (L)} \\
    \midrule
    Apple & Pseudo-RGB: [598, 548, 449]            & 0.176 & 0.225 \\
          & Bruise-contrast: [801, 708, 551]       & 0.118 & 0.245 \\
    \midrule
    Maize & Visible: [650, 540, 470]               & 0.147920 & 0.395930 \\
          & NIR-inclusive: [850, 580, 490]          & 0.147920 & 0.395930 \\
    \midrule
    Pear  & Pseudo-RGB                             & 0.0881612 & 0.125526 \\
          & Disease-oriented: [550, 710, 980]      & 0.0881611 & 0.125525 \\
    \bottomrule
    \end{tabular}
\end{table}

\section{Conclusions}
\label{sec:conclusions}

HSI-SC-NeRF is a stationary-camera hyperspectral NeRF framework tailored for agricultural applications requiring joint 3D structural and spectral analysis. Fixed-camera hyperspectral imaging combined with object rotation and simulated pose estimation eliminates the need for complex moving-camera rigs while preserving high-resolution geometry and spectral detail. The framework achieves near-perfect geometric reconstruction (F-score: 97.31\%) and captures spectrally meaningful information across the visible and near-infrared spectrum, with the ablation study confirming that a magnitude-dominant hybrid loss with masked fine-tuning yields the best wavelength-resolved fidelity.

Several limitations remain. Spectral accuracy is sensitive to non-uniform illumination, as shadows or lamp drift can introduce reflectance inconsistencies; future work could address this using decomposition techniques such as Nerfactor~\citep{zhang2021nerfactor}. The fixed-camera, rotating-object design simplifies hardware but may limit acquisition speed and requires careful calibration to avoid misalignment. The method also assumes rigid-body motion, which may not hold for soft or deformable objects, and reliance on simulated poses may reduce generalization to more complex or outdoor environments.

The high-fidelity 3D hyperspectral reconstructions produced by HSI-SC-NeRF capture detailed surface geometry and spectral characteristics critical for detecting subtle physiological traits such as early produce damage or stress. For postharvest supply chains, the framework enhances detection of defects including bruising, pathogen infestations, and physiological disorders, reducing waste and improving marketability~\citep{song2024fruit, li2025bruise}. For breeding programs and phenotyping initiatives, it enables non-destructive evaluation of complex traits that combine structural and biochemical components, accelerating selection cycles.

\section*{Acknowledgements}
This work was supported by the AI Institute for Resilient Agriculture (USDA-NIFA 2021-67021-35329), NSF 2412929/2412928 and Iowa State University's Plant Science Institute. 

\bibliographystyle{elsarticle-num-names} 
\bibliography{CEARefs}

\appendix

\section{Automatic central mask generation}
\label{sec:app_wr_auto_mask}

To ensure robust spectral calibration, we performed a spatial deviation analysis on the white reference (WR) tarp to automatically identify a uniform and illumination-stable calibration region. A per-pixel relative deviation map was computed by comparing each WR pixel to the band-wise mean WR spectrum, which revealed increased deviation near the tarp edges due to illumination falloff and boundary effects.

\begin{figure*}[t!]
    \centering
    \includegraphics[width=0.84\textwidth]{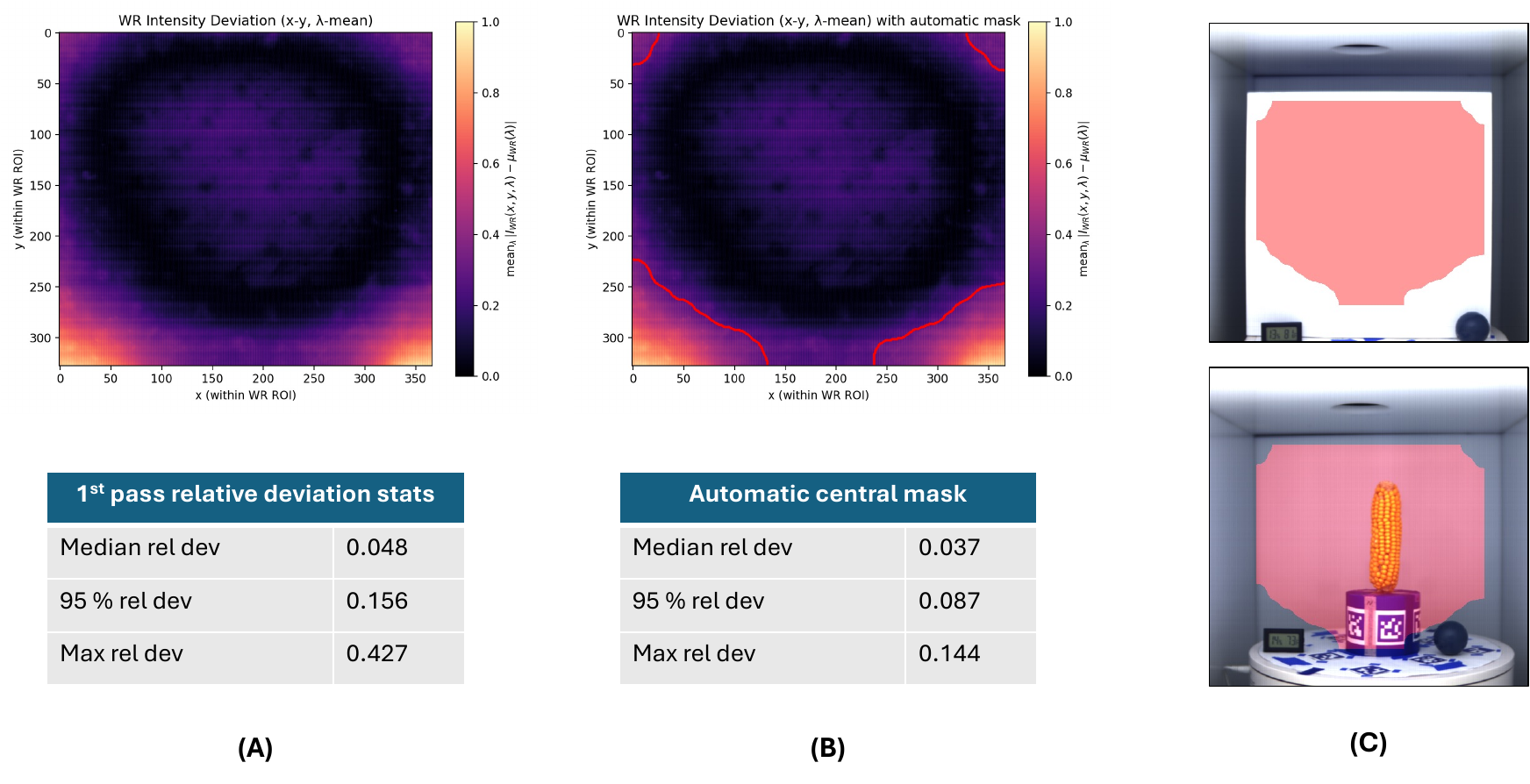} 
    \caption{Automatic central WR mask generated using the 70th-percentile deviation threshold.
    (A) First-pass spatial relative deviation map computed over the WR region of interest.
    (B) Refined deviation map and resulting automatic central WR mask after percentile thresholding and morphological filtering.
    (C) Example acquisition showing the WR tarp and the object positioned at the center of the scene.}
    \label{fig:auto_mask_70}
\end{figure*}

\begin{figure*}[t!]
    \centering
    \includegraphics[width=0.84\textwidth]{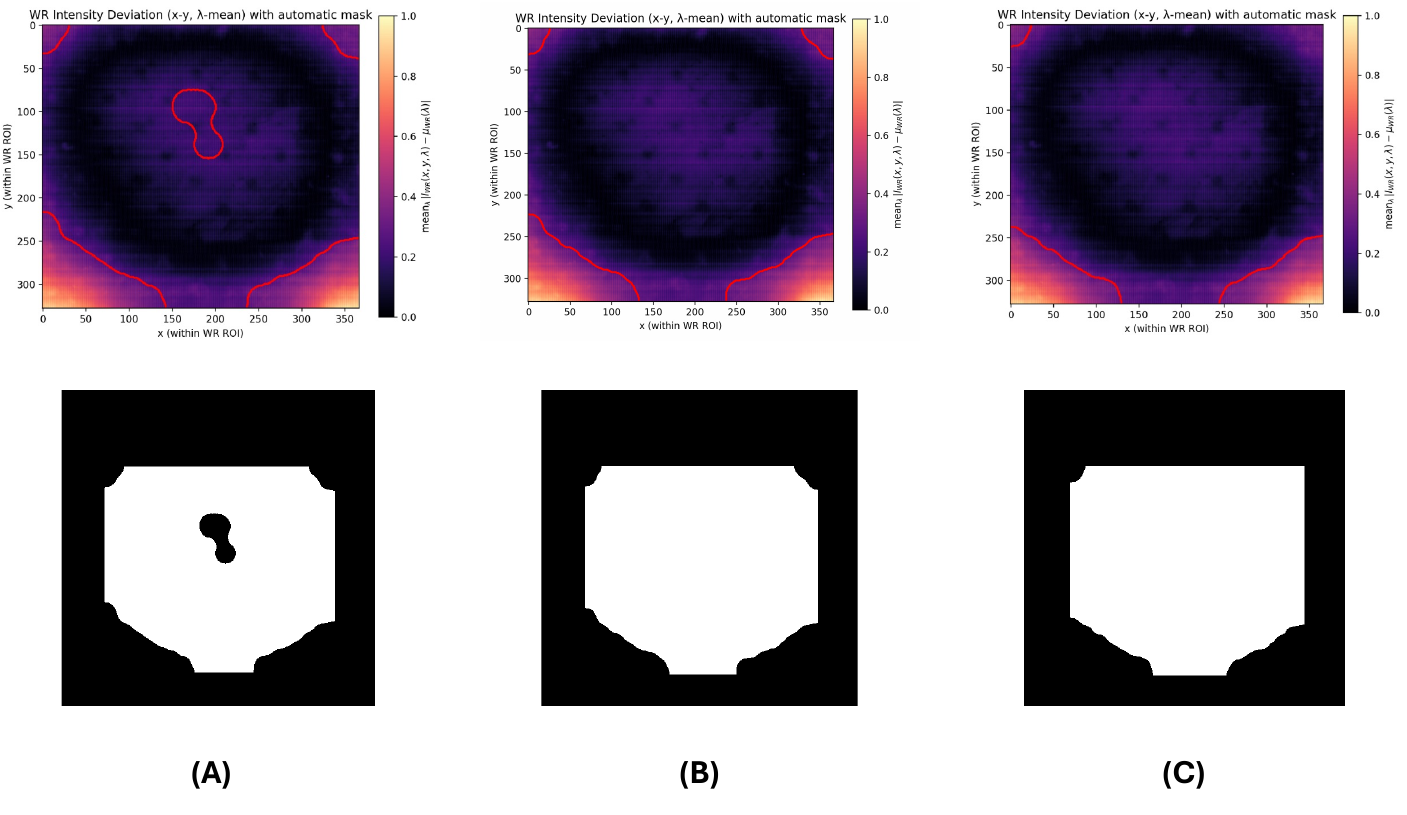}
    \caption{Justification of the 70th-percentile threshold for automatic WR mask generation. In each pair, the top plot shows the spatial relative deviation map computed within the WR region of interest, and the bottom image shows the corresponding binary WR mask projected onto the original image. (A), (B), and (C) correspond to the 65th-, 70th-, and 75th-percentile thresholds, respectively.}
    \label{fig:percentile_threshold}
\end{figure*}

As shown in \figref{fig:auto_mask_70}, applying a 70th-percentile threshold yields a spatially contiguous central WR mask that significantly reduces median, 95th-percentile, and maximum relative deviations compared to the unfiltered WR region, while retaining approximately \(1.05 \times 10^{5}\) pixels for calibration. \figref{fig:percentile_threshold} further illustrates the effect of varying the percentile threshold. Lower thresholds (e.g., 65\%) are overly restrictive and remove spatially valid WR pixels in the central region, resulting in holes within the mask, whereas higher thresholds (e.g., 75\%) begin to include edge-affected and illumination-unstable pixels. The 70th-percentile threshold therefore represents a practical trade-off in which deviation statistics stabilize and spatial coherence in the central region—where the object is located—is preserved. This selection reflects robustness considerations rather than hyperparameter tuning.

\end{document}